\documentclass[times,twocolumn,final,authoryear]{elsarticle}

\usepackage{ycviu}
\usepackage{framed,multirow}

\usepackage{amssymb}
\usepackage{latexsym}
\usepackage{xspace}
\usepackage{hyperref}
\usepackage{times}
\usepackage{epsfig}
\usepackage{graphicx}
\usepackage{amsmath}
\usepackage{booktabs}
\usepackage{float}
\usepackage{caption}
\usepackage{subcaption}
\usepackage{dsfont}
\usepackage{mathtools}
\usepackage{diagbox}

\usepackage{url}
\usepackage{xcolor}
\definecolor{newcolor}{rgb}{.8,.349,.1}

\newcommand{\methodname}{COLOSEO\xspace}
\newcommand{\baseline}{COLOSEO-$\mathcal{E}$\xspace}
\newcommand{\task}{OUVDA\xspace}
\newcommand{\data}{\mathcal{D}}
\newcommand{\tsource}{\mathtt{S}}
\newcommand{\ttarget}{\mathtt{T}}
\newcommand{\R}{\mathbb{R}}

\newcommand{\bfh}{{\bf h}}

\newcommand{\bfu}{{\bf u}}
\newcommand{\bfv}{{\bf v}}

\newcommand{\bfx}{{\bf x}}

\newcommand{\bfz}{{\bf z}}

\newcommand{\bfX}{{\bf X}}

\newcommand{\calB}{{\mathcal B}}

\newcommand{\calL}{{\mathcal L}}

\newcommand{\calX}{{\mathcal X}}
\newcommand{\calY}{{\mathcal Y}}

\def\ie{\emph{i.e.}}
\def\eg{\emph{e.g.}}
\def\etc{\emph{etc.}}

\newcommand{\ths}{\textsuperscript{th}\;}

\journal{Computer Vision and Image Understanding}

\begin{document}

\thispagestyle{empty}

\begin{frontmatter}

\title{Simplifying Open-Set Video Domain Adaptation with Contrastive Learning}

\author[1]{Giacomo \snm{Zara}
} 
\author[1]{Victor \snm{Guilherme Turrisi da Costa}}
\author[2]{Subhankar \snm{Roy}}
\author[1]{Paolo \snm{Rota}}
\author[1,2]{Elisa \snm{Ricci}}

\address[1]{University of Trento, Via Sommarive 9, Trento, Italy}
\address[2]{Fondazione Bruno Kessler, Via Sommarive 18, Trento, Italy}

\received{1 May 2013}
\finalform{10 May 2013}
\accepted{13 May 2013}
\availableonline{15 May 2013}
\communicated{S. Sarkar}

\begin{abstract}
In an effort to reduce annotation costs in action recognition, unsupervised video domain adaptation methods have been proposed that aim to adapt a predictive model from a labelled dataset (i.e., source domain) to an unlabelled dataset (i.e., target domain). In this work we address a more realistic scenario, called open-set video domain adaptation (OUVDA), where the target dataset contains ``unknown'' semantic categories that are not shared with the source. The challenge lies in aligning the shared classes of the two domains while separating the shared classes from the unknown ones. In this work we propose to address OUVDA with an unified contrastive learning framework that learns discriminative and well-clustered features. We also propose a video-oriented temporal contrastive loss that enables our method to better cluster the feature space by exploiting the freely available temporal information in video data. We show that discriminative feature space facilitates better separation of the unknown classes, and thereby allows us to use a simple similarity based score to identify them. We conduct thorough experimental evaluation on multiple OUVDA benchmarks and show the effectiveness of our proposed method against the prior art.
\end{abstract}


\end{frontmatter}



\section{Introduction}
\label{sec:introduction}

Action recognition is an important problem in the field of computer vision where the task consists in recognizing the action being performed in a video sequence. Supervised action recognition~\citep{tran2015learning,feichtenhofer2016convolutional,kinetics_i3d,zhou2018temporal} is widely studied because of the growing need for automatically categorizing video content that are being generated everyday. However, it is nearly impossible for human annotators to keep pace with the enormous volumes of online videos, and thus supervised training becomes infeasible. A cheaper way of leveraging the massive pool of unlabelled data is by exploiting an already trained model to infer the labels on such data and then re-using them to build an improved model. Such an approach is also prone to failure because the unlabelled data may belong to a data distribution that is different from the annotated one, which is often referred to as the \textit{domain-shift} problem~\citep{torralba2011unbiased}.


To address the domain-shift problem, \textit{Unsupervised Video Domain Adaptation} (UVDA) methods~\citep{ta3n,sava,tcon} have been proposed that aim to learn a model for the domain of interest by jointly leveraging the annotated \textit{source} data and the unannotated \textit{target} data. However, these methods make a strong and unrealistic assumption that the source and target domains share the same label space, which is also known as the \textit{closed-set} scenario. The closed-set assumption is rarely the case in 
the real world as the target domain can contain video sequences from action categories that are not present in the source dataset. Such non-overlapping categories are known as the \textit{out-of-distribution} (OOD) classes~\citep{vaze2021open}. A naive application of the existing UVDA techniques will cause the model to incorrectly classify the OOD classes as one of the shared ones, which is undesirable.

Due to the limited applicability of the closed-set UVDA setting, the focus has been shifting towards the more challenging \textit{open-set} scenario where the target dataset contains samples associated with categories that are not present in the source domain. In the context of action recognition, this task is referred to as the \textit{Open-Set Unsupervised Video Domain Adaptation} (\task)~\citep{cevt}. The main goal in \task consists in adapting a model to the target domain that can align the classes that are shared between the two domains while excluding the OOD (also known as \textit{target-private} or \textit{unknown}) classes from the alignment process. Although open-set unsupervised domain adaptation has received significant attention for the image classification task~\citep{Saito_2018_ECCV,dance,hscore,ovanet,bucci}, it is rather understudied in the field of action recognition~\citep{cevt}.

\begin{figure}
    \centering
    \includegraphics[width=\columnwidth]{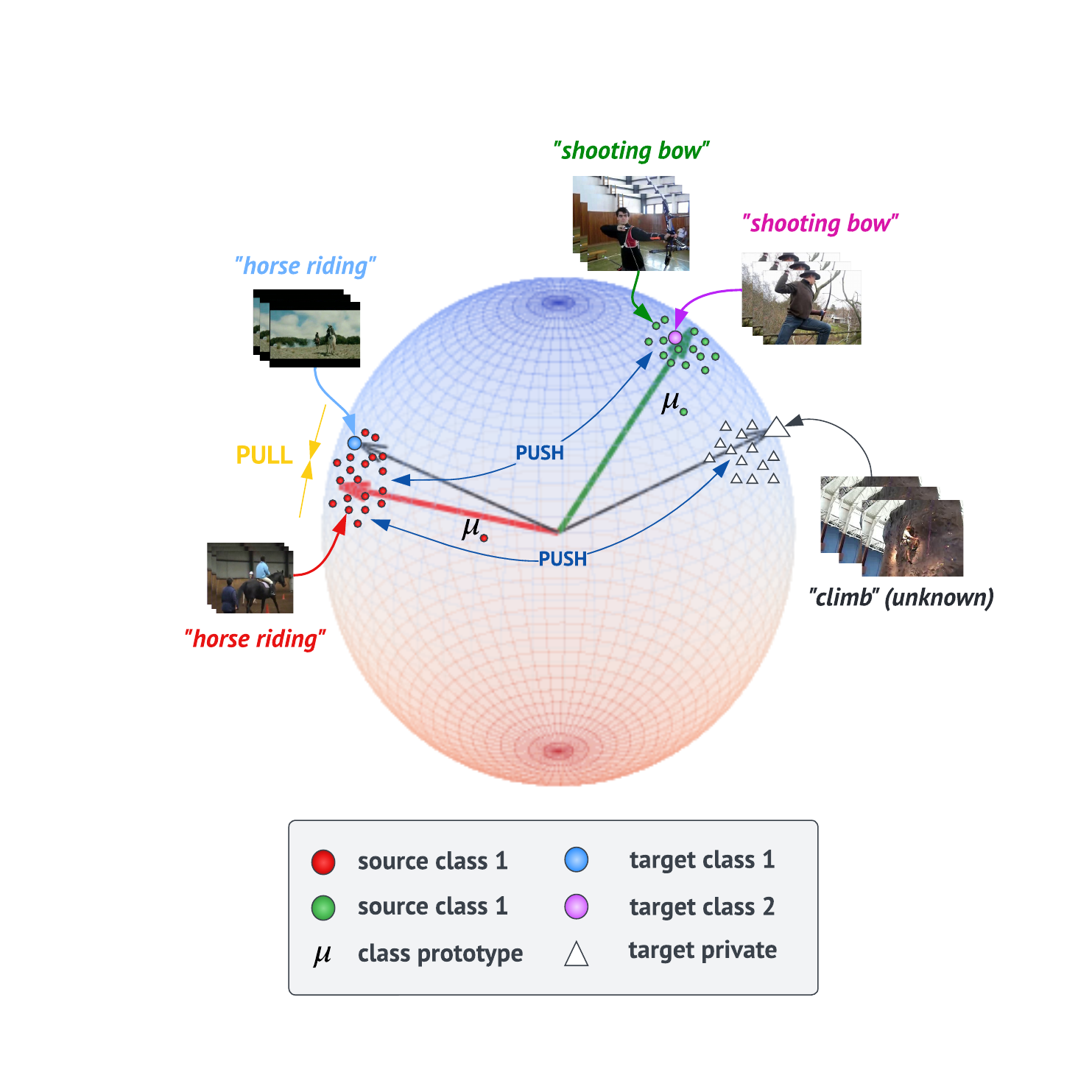}
    \caption{An illustration of a feature space learned with contrastive learning in order to obtain discriminative video features. A well-clustered feature space simplifies the separation of the shared classes from the out-of-distribution (or target-private) ones.}
    \label{fig:teaser}
\end{figure}

The recently proposed \task method, CEVT~\citep{cevt}, uses a weighted adversarial learning strategy, with the weights derived from class-conditional extreme value theory, in order to recognize the target-private classes. However, adversarial learning-based alignment can be unstable in practice, especially in the open-set scenarios. In this work, we argue that \task can be greatly simplified if we can learn \textit{discriminative features} on both the source and target data for the following two reasons: (i) it can implicitly align the shared classes between the source and target domains by learning semantically distinct clusters~\citep{wang2020understanding}; and (ii) it results in the target-private classes being well separated from the shared ones (see Fig.~\ref{fig:teaser}). Based on this intuition, we propose to leverage contrastive learning~\citep{gutmann2010noise,simclr,moco} to obtain discriminative representations to tackle \task.

In details, we realize the above goals by proposing to use different instantiations of the contrastive loss that mainly differ in how the \textit{positives} and \textit{negatives} are mined. To recap, the contrastive loss~\citep{simclr} relies on positives and negatives which are then contrasted to learn feature representation. In our proposed method we use four such instantiations of contrastive loss: (i) a \textit{label}-based supervised contrastive loss for the source; (ii) an \textit{augmentation}-based contrastive loss for the target; (iii) a \textit{cross-domain} contrastive loss between the source and the pseudo-labelled target; and (iv) a \textit{temporal} contrastive loss to learn the temporal dynamics present in video data. One important advantage with such a proposal is that we can leverage an unique contrastive formulation to \textit{unify} several losses, which cater to different learning aspects of the task at hand. It also ensures compatible gradients and dispose of the need of tuning several hyperparameters. To this end, we call our proposed method {\bf \methodname} (\textbf{CO}ntrastive \textbf{L}earning for \textbf{O}pen-\textbf{SE}t Vide\textbf{O} Domain Adaptation).

Owing to the well-clustered feature space we observe that it becomes fairly straightforward to separate the target-private instances from the shared ones. In this work we use the nearest class prototypes~\citep{mensink2013distance} with a simple cosine-similarity metric to exclude the target-private instances from the alignment process that lie far from all the source classes prototypes. We extensively validate our proposed method on multiple benchmarks: (i) \textit{HMDB$\leftrightarrow$UCF} and \textit{UCF$\leftrightarrow$Olympic} that contains actions from third-person view; and (ii) the challenging first-person (egocentric) \textit{Epic-Kitchens}.

\paragraph{\bf Contributions} To summarize, our main contributions are: (i) the {\bf \methodname} framework, which uses an unified contrastive learning formulation to address the task of \task. We demonstrate that learning compact and discriminative feature representations can greatly simplify the \task task; (ii) a novel video-oriented temporal contrastive loss that improves the OOD robustness, thereby facilitating the task at hand; (iii) we advance the state-of-the-art on three standard \task benchmarks by non-trivial margins.


\section{Related Works}
\label{sec:related_work}



\paragraph{\bf Open-set Unsupervised Domain Adaptation}
There exists a large body of the open-set unsupervised domain adaptation (OUDA) literature but applied specifically to the image classification task. OUDA was first introduced by \cite{Busto_2017_ICCV} and later then addressed in several subsequent works, which mostly rely either on adversarial learning or clustering approaches.

Most of the initial works~\citep{Saito_2018_ECCV, 9165955} exploited adversarial learning to detect target-private samples while aligning features of different domains for the known categories. \cite{8803287} improved the adversarial objective of \cite{Saito_2018_ECCV} by replacing the binary cross entropy loss with a symmetrical Kullback-Leibler distance-based loss. \cite{Liu_2019_CVPR} addressed OUDA by training a multi-binary classifier with source data, alongside domain discriminator, to progressively separate the samples of unknown and known classes. A graph neural network based approach was proposed by \cite{pmlr-v119-luo20b} that integrates episodic pseudo-labelling with adversarial learning to tackle the OUDA task. 

Recently, the adversarial approaches have been replaced by clustering-based methods~\citep{9316151} that ensure well clustered target space. For instance, \cite{bucci} proposed HyMOS that uses contrastive learning to enforce invariance to the augmentations obtained via style transfer. However, HyMOS is tailored for the multi-source OUDA setting. \cite{Ma_2021_ICCV} introduced a method for active OUDA that combines adversarial learning with the clustering non-transferable gradient embedding approach. Recently, \cite{dance} proposed DANCE that uses self-supervision to cluster the target data, followed by detecting the target-private instances based on the entropy computed by the classifier. Different to all other approaches, OVANet~\citep{ovanet} trains one-vs-all binary open set classifiers on the source data to detect and reject the unknown samples. A common theme in all these above methods is that they are have been proposed for images and do not exploit the temporal information which video data has to offer.

\paragraph{\bf Open-set Unsupervised Video Domain Adaptation}
In the realm of action recognition the open-set video domain adaptation (\task) the literature is relatively under explored. \cite{busto2019open} proposed a generic approach based on learning a mapping from the source to the target domain and learning an open-set classifier on the mapped samples. CEVT was introduced by \cite{cevt} that models the entropy of the target samples as generalised extreme value distribution in order to perform open-set separation. Differently from the prior art CEVT, our proposed approach additionally exploits the temporal dynamics in the video sequences. We observe that modelling the temporal dynamics can lead to better representations, which leads to even compact action clusters. Moreover, the adopted contrastive learning objective allows us to align the two domains without the need of any adversarial feature alignment.

\paragraph{\bf Contrastive Representation Learning}
As collecting annotated data is costly, the research community has focused on self-supervised representation learning (SSL) that learns discriminative features unsupervisedly with the help of a pretext task. Contrastive learning~\citep{gutmann2010noise,simclr,moco} is one such family of SSL methods that casts the learning as an instance discrimination task where the similarity between two correlated views are maximized. So far in the literature, contrastive learning has recently been exploited in an attempt to learn discriminative features, but only in the context of closed-set UVDA~\citep{ek_multimodal,kim2021learning, cont_mix,9707012,costa2022unsupervised}. Moreover, contrastive learning has found to be effective for detecting OOD images samples~\citep{winkens2020contrastive}. Different from these works, our proposed formulation of different contrastive losses takes into account the temporal dimension in video. In our experiments we show that the proposed temporal contrastive loss is influential in improving over the image-based counterparts.
\section{Methods}
\label{sec:method}

In this work, we propose {\bf \methodname} for addressing the Open-set Unsupervised Video Domain Adaptation (\task) task. We first formally define \task and then we introduce our proposed method.

\paragraph{\bf Problem Definition and Notations}
\label{subsec:problem}

Let us assume that there is a labelled source dataset $\data^\tsource = \{(\bfX^\tsource_i, y^\tsource_i)\}_{i=1}^{N^\tsource}$ containing $N^\tsource$ instances, where $\bfX \in \calX$ represents an input and $y \in \calY$ the corresponding $K$ semantic categories. We also assume that there is an unlabelled target dataset $\data^\ttarget = \{\bfX^\ttarget_i\}^{N^\ttarget}_{i=1}$ of $N^\ttarget$ instances, containing the $K$ \textit{shared} semantic categories of the source dataset $\data^\tsource$ and some additional categories, denoted as \textit{target-private} or \textit{unknown} classes\footnote{In this work target-private and unknown classes are used interchangeably}, which are absent in the $\data^\tsource$. These unknown classes are designated as the $(K+1)$\ths class. As per standard unsupervised domain adaptation assumption, the underlying marginal probability distributions between the source and the target domains differ from each other, \ie, $p(\bfX^\tsource) \neq p(\bfX^\ttarget)$. Under such conditions, the goal in the \task is to learn a mapping $f_\theta \colon \calX \to \calY$, modelled by a neural network $f$ with parameters \(\theta\), that can correctly classify the shared instances of the $\data^\ttarget$ into the first $K$ classes and all the target-private instances into the $(K+1)$\ths class.

Since we are working with arbitrary-length video sequences, during training, we sub-sample frames from each video sequence to form constant-length \textit{clips}. More formally, each input sample $\bfX \in \calX$ is given as $\bfX = \{\bfx_k\}^{M}_{k=1}$, where $M$ denotes the number of frames used in constructing the clip. As it is done typically in the action recognition pipelines, $c$ clips from a video sequence are presented as input to $f_\theta$ and then later summarized (or aggregated) to provide the final class prediction.


\begin{figure*}[t!]
    \centering
    \includegraphics[width=\textwidth]{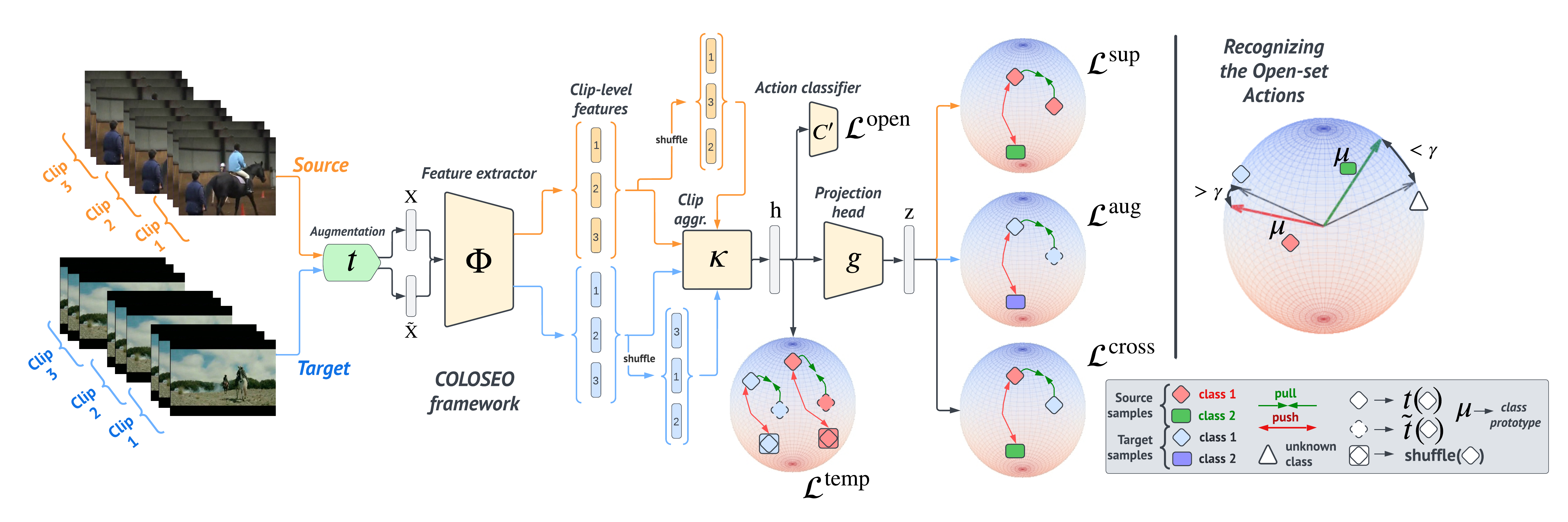}
    \caption{Overview of our proposed \methodname framework. \textbf{Left}: The video sequences are divided into clips, which are input to the feature extractor $\Phi$ to obtain clip-level features. The clip-level features are then aggregated into the final video-level features $\bfh$ by the clip aggregator network $\kappa(\cdot)$. We have a projection head $g(\cdot)$ that projects the features $\bfh$ to a hypersphere, producing $\bfz$. The four instantiations of the contrastive losses: $\mathcal{L}^\text{sup}$, $\mathcal{L}^\text{aug}$, $\mathcal{L}^\text{cross}$ and $\mathcal{L}^{\text{temp}}$ are used to learn compact and discriminative features. A well-clustered feature space enables easy separation between the shared and unknown classes. The $\calL^\text{open}$ is used to learn a $(K+1)$-way classifier $C'$ for the target samples. \textbf{Right}: A cosine similarity-based nearest class-prototype metric determines if a target sample belongs to the shared classes or not, where $\gamma$ is used as threshold.}
    \label{fig:method_overview}
\end{figure*}

\paragraph{\bf Overview}

In this work we propose to simplify the \task task by resorting to \textit{contrastive learning}~\citep{simclr, moco}. We choose contrastive learning because it learns very discriminative representations that are often useful for several downstream tasks~\citep{ericsson2021self}. Moreover, it has also been shown in the work by~\cite{dance} that learning discriminative representations greatly alleviates the misalignment problem between the shared and unknown classes owing to the well-structured feature space. Furthermore, self-supervised representation learning has been found useful for detecting OOD samples~\citep{hendrycks2019using}, which is also a key constituent challenge in the \task.


Our proposed \methodname is divided into two stages. In the first stage, we aim at learning useful video-level feature representations on both the source and the target data such that instances with the same underlying action are clustered together, facilitating the adaptation phase. Since the source data is labelled we use both the standard cross-entropy classification loss and the supervised contrastive loss (Sup-Con)~\citep{khosla2021supervised}, such that instances from the same class are pulled closer to each other, while instances from different class are pushed apart. As the target data lack labels, we adopt the augmentation-based SimCLR~\citep{simclr} loss to cluster the instances in the target domain.
Lastly, as temporal dynamics are crucial for distinguishing between different actions, we propose a \textit{temporal contrastive} loss that contrasts between non-shuffled and shuffled clips in a given video sequence. The temporal contrastive loss allows the network to learn the salient temporal dynamics in an action, which is important for the recognition of the action (see Fig.~\ref{fig:method_overview} left).

The second stage deals with further aligning the feature representations of the source and target domains, while ensuring that the target-private instances are not aligned with the shared classes. To this end, we introduce a simple score-based nearest prototype classifier which determines if a target sample belongs to the shared classes or not. This score is measured with respect to the source prototypes (or class centroids) in order to be less prone to noisy outliers (see Fig.~\ref{fig:method_overview} right). If a target sample is deemed to belong to one of the shared classes, we find its pseudo-label with the help of the closed-set source classifier and employ the Sup-Con loss to align the source and the target representations. On the other hand, if the target sample is detected to be far from all the source prototypes, it is then excluded from the cross-domain feature alignment. A $(K+1)$-way classifier is learned in the second stage that additionally classifies the detected target-private instances into the $(K+1)$\ths class. An overview of our method is provided in the Fig.~\ref{fig:method_overview}, and we describe it in detail below.

\subsection{COLOSEO: Contrastive Learning for \task}
\label{subsec:simclr}

To address the \task, we employ contrastive learning~\citep{simclr, moco} in both the stages to (i) produce discriminative feature representations for both the source and target data; and (ii) align the source and target domains over the shared classes. We realize the above goals by using four different versions of the contrastive losses that differ in a way the \textit{positives} and \textit{negatives} are created, which are: (i) the label-based Sup-Con~\citep{khosla2021supervised} $\calL^\text{sup}$, where positives consist of samples with the same class label and negatives are the samples with a different class label; (ii) the augmentation-based contrastive loss $\calL^\text{aug}$ where the positives are the augmented views of a given sample, and negatives consist of all other samples in the mini-batch; (iii) the temporal contrastive loss $\calL^\text{temp}$ where the positives are the same as in the augmentation-based $\calL^\text{aug}$ loss, but the negatives are created by shuffling the order of the clips of the same video; and (iv) the cross-domain contrastive loss $\calL^\text{cross}$ which is similar to the $\calL^\text{sup}$, except that the pseudo-labels are utilized for the unlabelled target samples.


Given a mini-batch of source domain video sequences $\calB^\tsource=\{(\bfX^\tsource_i, y^\tsource_i)\}_{i=1}^{b}$ and target domain video sequences $\calB^\ttarget={\{\bfX^\ttarget_i\}_{i=1}^{b}}$ of size $b$, we apply two stochastic data augmentation transformations $t(\cdot)$ and $\tilde{t}(\cdot)$ on each sample to create the positive pairs (or \textit{views}) as $(\bfX^\tsource_i$, $\tilde{\bfX}^\tsource_i)$ and $(\bfX^\ttarget_i$, $\tilde{\bfX}^\ttarget_i)$. As each video sequence $\bfX$ consists of $c$ clips, we first forward the clips through the feature extractor $\Phi(\cdot)$ to obtain $c$ clip-level features, which are then fused using the clip aggregator network $\kappa(\cdot)$. In detail, for every input video sequence $\bfX_i$ the aggregator network $\kappa(\cdot)$ yields a video-level feature $\bfh_i = \kappa(\Phi(\bfX_i)) \in \R^{1024}$. Thus, corresponding to the positive pairs $(\bfX^\tsource_i$, $\tilde{\bfX}^\tsource_i)$ and $(\bfX^\ttarget_i$, $\tilde{\bfX}^\ttarget_i)$ in the input space we have $(\bfh^\tsource_i, \tilde{\bfh}^\tsource)$ and $(\bfh^\ttarget, \tilde{\bfh}^\ttarget)$ in the video-level feature space $\R^{1024}$. Following, the previous works on self-supervised learning~\citep{simclr,khosla2021supervised}, we also use a non-linear \textit{projection head} $g(\cdot)$ that operates on the video-level features producing $\bfz = g(\bfh)$.  


\paragraph{\bf Label-based Contrastive Loss} Since the source data is annotated, we exploit the class label information to create more informative positives and negatives. In detail, given an instance $\bfX^\tsource_i$ belonging to the class $y^\tsource_i$, another instance $\bfX^\tsource_j$ is a positive in the mini-batch if it shares the same class label, \ie, $y^\tsource_i = y^\tsource_j$. Similarly, all other instances in the mini-batch which do not share the same class label with $\bfX^\tsource_i$ are considered negatives. The label-based contrastive loss is then defined, for the $i$\ths source sample, as:



\begin{flalign}
\calL^{\text{sup}}_{i} = - \log \sum_{j=1}^{2b} \frac{\mathds{1}_{y^\tsource_i = y^{\tsource}_j} \exp(\frac{\text{sim}(\bar{\bfz}^{\tsource}_i, \bar{\bfz}^{\tsource}_j)}{\tau})}{  \splitfrac{ \exp(\frac{\text{sim}(\bar{\bfz}^{\tsource}_i, \bar{\bfz}^{\tsource}_j)
}{\tau}) +}{ \displaystyle \sum_{k=1}^{2b} \mathds{1}_{k \neq i} \mathds{1}_{y^{\tsource}_k \neq y^{\tsource}_i} \exp(\frac{\text{sim}(\bar{\bfz}^{\tsource}_i, \bar{\bfz}^{\tsource}_k)}{\tau})  } },
\label{eq:loss_sup_con}
\end{flalign}

where $\bf\bar{z}^{\tsource} = \bfz^{\tsource} \cup \bf\tilde{z}^{\tsource}$, $\text{sim}(\bfu, \bfv) =\frac{\bfu^{\intercal} \bfv}{\Vert \bfu \Vert \Vert \bfv \Vert}$ is the dot product between the L2 normalized $\bfu$ and $\bfv$ and $\tau$ is a temperature parameter. The operator $\mathds{1}_{k \neq i}$ evaluates to 1 with $k \neq i$ or 0 otherwise. Similarly, the operator $\mathds{1}_{y^{\tsource}_k \neq y^{\tsource}_i}$ yields 1 if the class labels are different, or 0 otherwise.



\paragraph{\bf Augmentation-based Contrastive Loss} For the unlabelled target data, we use the augmentation-based contrastive loss. Given an instance $i$, its positive is defined as the augmented view of itself and its negatives are all the other instances in the mini-batch. We can define the \textit{augmentation-based contrastive} loss on the $i$\ths target sample as:

\begin{equation}
\calL^{\text{aug}}_i = - \log \frac{\exp(\frac{\text{sim}(\bfz^{\ttarget}_i, \tilde{\bfz}^{\ttarget}_i)}{\tau})}{\sum_{k=1}^{2b} \mathds{1}_{k \neq i} \exp(\frac{\text{sim}(\bfz^{\ttarget}_i, \bfz^{\ttarget}_k)}{\tau})}.
\label{eq:loss_nce}
\end{equation}

We make the $\calL^{\text{aug}}$ loss symmetric by swapping $\bfz^\ttarget$ and $\tilde{\bfz}^\ttarget$ and averaging the losses.

\paragraph{\bf Temporal Contrastive Loss} While the self-supervised contrastive loss is very much potent at modelling the shapes of objects in the clips while ignoring the low-level appearance information, it still might not be sufficient in the case of action recognition due to the temporal dynamics associated with an action. Consider an example, where the network must discriminate between the actions ``\texttt{push-up}'' and ``\texttt{pull-up}''. Since both these classes depict humans performing work-out routines, the invariances induced by the strong augmentations in the contrastive loss would favour discovering the \textit{human shape}, which is confounding for the two action classes. However, the classes ``\texttt{push-up}'' and ``\texttt{pull-up}'' inherently exhibit different temporal dynamics, which are unique of their own. We conjecture that, asides from modelling the shape with the contrastive loss, capturing the temporal dynamics can lead to even more discriminative features. Thus, in order to benefit from the temporal information inherently conveyed in video data, we design a video-oriented \textit{temporal contrastive} loss that contrasts between video sequences with shuffled and non-shuffled clips.

\begin{figure}[!t]
    \centering
    \includegraphics[width=0.8\columnwidth]{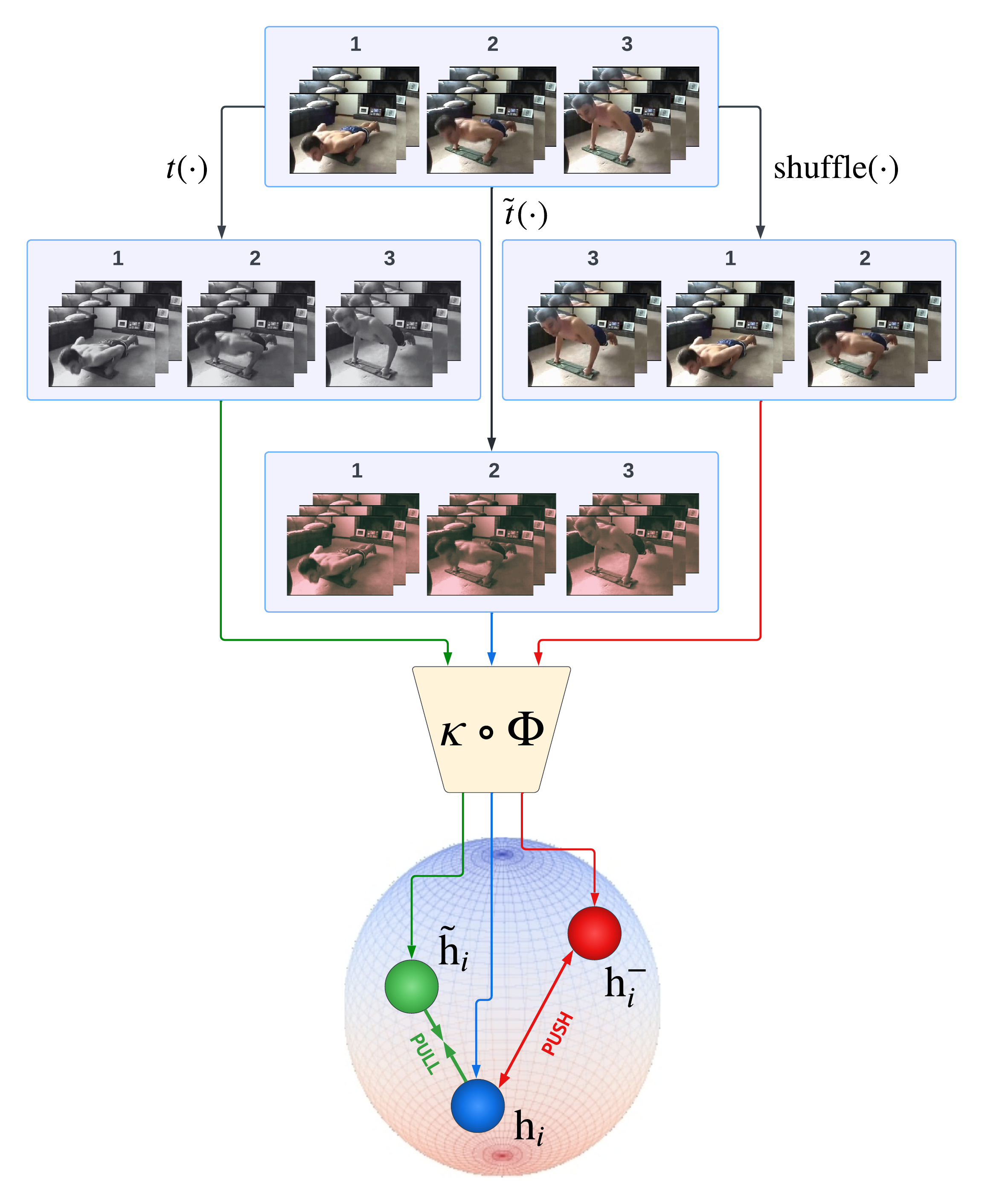}
    \caption{An overview of the temporal contrastive loss \(\mathcal{L}^\text{temp}\). The anchor and positive are created with transformations $t(\cdot)$ and $\tilde{t}(\cdot)$, respectively. The negatives are artificially generated by shuffling the clips. The \(\mathcal{L}^\text{temp}\) learns the temporal dynamics by pulling the anchor towards the positive, and pushing the negative far from the anchor}
    \label{fig:triplet_loss}
\end{figure}

As before, given a sample $\bfX_i$, we create its positive view $\tilde{\bfX}_i$ with the help of a strong augmentation. Whereas, for contrasting with a regular video sequence, we artificially generate negatives by randomly permuting the order of clips in $\bfX_i$, producing $\bfX^-_i$. However, as the contrastive loss requires a significant amount of negatives in a mini-batch to work~\citep{simclr,moco}, one would need perform repeated forward passes through the clip aggregator network for every permutation. To circumvent this, we simplified the contrastive formulation to a triplet loss~\citep{chechik2010large}, which achieves a similar effect. Then, the goal is to push the anchor $\bfX_i$ close to the augmented sequence $\tilde{\bfX}_i$, while pushing $\bfX_i$ away from the \textit{shuffled} $\bfX^-_i$ (see Fig.~\ref{fig:triplet_loss}) in the feature space. The temporal contrastive loss for the $i$\ths sample is given as:

\begin{equation}
    \label{eq:triplet_loss}
    \calL^{\text{temp}}_i = \max\{d(\bfh_i, \tilde{\bfh}_i) - d(\bfh_i, \bfh^-_i) + \alpha, 0\}
\end{equation}
where $d(\cdot, \cdot)$ is a distance function defined as \(d(\cdot, \cdot) = ||\cdot - \cdot||_2\), where \(\alpha\) indicates the margin of the triplet loss.

\paragraph{\bf Cross-domain Contrastive Loss} While the aforementioned contrastive losses can lead to a well-clustered feature space, they do not guarantee that the source and target class clusters are perfectly aligned. This is primarily caused by domain shift. To further align the two domains, we propose to use a \textit{cross-domain contrastive} loss that is similar to the label-based contrastive loss $\calL^{\text{sup}}$, except that pseudo-labels are used for the unlabelled target samples. In particular, to align a target sample $\bfX^{\ttarget}_i$ with the source samples, we first infer the pseudo-label $\hat{y}^{\ttarget}_i$ of the $i$\ths target sample using the source classifier $C(\cdot)$. Then we proceed as in Eq.~(\ref{eq:loss_sup_con}) to define the cross-domain contrastive loss as:

\begin{flalign}
\calL^{\text{cross}}_{i} = - \log \sum_{j=1}^{2b} \frac{\mathds{1}_{\hat{y}^\ttarget_i = y^{\tsource}_j} \exp(\frac{\text{sim}(\bfz^{\ttarget}_i, \bfz^{\tsource}_j)}{\tau})}{  \splitfrac{ \exp(\frac{\text{sim}(\bfz^{\ttarget}_i, \bfz^{\tsource}_j)
}{\tau}) +}{ \displaystyle \sum_{k=1}^{2b} \mathds{1}_{k \neq i} \mathds{1}_{y^{\tsource}_k \neq \hat{y}^{\ttarget}_i} \exp(\frac{\text{sim}(\bfz^{\ttarget}_i, \bfz^{\tsource}_k)}{\tau})  } },
\label{eq:loss_cross_domain}
\end{flalign}

While inferring the pseudo-label of a target sample with the $K$-way source classifier $C(\cdot)$ may work seamlessly in the closed-set UVDA, it is problematic in the \task case because of the presence of the target-private classes. To prevent the target-private classes from being assigned to one of the shared categories, we present an unknown class detection protocol that eliminates the target samples that are unlikely to belong to one of the shared classes. Next we elaborate the unknown class detection protocol.

\subsection{Recognizing the Open-set Actions}
\label{subsec:open_set_protocol}

Before the pseudo-labels are inferred for the target samples, we need a mechanism that can automatically detect the target-private classes. For this purpose, we design a metric which assigns a \textit{score} to each target sample. A low score means that the target sample's representation is far from the training data, and is likely to be a target-private instance. For this purpose, we use \textit{cosine similarity} as the metric between the target feature $\bfh^\ttarget_i$ and the nearest source \textit{class prototype}. The class prototype $\boldsymbol{\mu}_\mathnormal{k}$ for a source class $\mathnormal{k}$ is obtained by averaging the features of all the training samples of class $\mathnormal{k}$, which is defined as:

\begin{equation}
\label{eq:prototypes}
\boldsymbol{\mu}_\mathnormal{k} = \frac{1}{N^\tsource_\mathnormal{k}} \displaystyle \sum^{N^\tsource_\mathnormal{k}}_{i=1} \kappa(\Phi(\bfX^\tsource_i))
\end{equation}
where $N^\tsource_\mathnormal{k}$ is the number of source samples from the class $\mathnormal{k}$. If the target feature $\bfh^\ttarget_i$ is farther than its nearest source class prototype $\boldsymbol{\mu}_\mathnormal{k}$ by a threshold $\gamma$, i.e., it has cosine similarity lower than the $\gamma$, we consider this sample as target-private. Thus, we have an indicator variable $s$ which is 1 if a target sample $\bfX^\ttarget_i$ is target-private and 0 otherwise. We can define it formally as:

\begin{equation}
\label{eqn:open-set-separation}
    s_i = \mathds{1}\left[ \text{max}_{\mathnormal{k} \in \{1, 2, \dots, K\}} (\text{sim}(\bfh^\ttarget_i, \boldsymbol{\mu}_\mathnormal{k})) \leq \gamma \right]
\end{equation}
where all the target samples with $s=1$ are \textit{not} included in the cross-domain contrastive loss $\calL^\text{cross}$.

Since the final goal in the \task is not only to classify the shared target instances into one of the $K$ classes but also to classify all the target-private instances into the $(K+1)$\ths class, we initialize a $(K+1)$-way classifier $C'(\cdot)$ in the second stage of training. All the target samples where $s=1$ using Eq.~(\ref{eqn:open-set-separation}) are assigned an ``unknown'' label $K+1$. We then use a standard cross-entropy loss to assign a high likelihood to the target private instances to be belonging to the $(K+1)$\ths class. Whereas, for the source samples we keep the standard classification loss. The cross-entropy loss is for a source $\bfX^\tsource_i$ and target-private instance $\bfX^\ttarget_{i, s=1}$ is defined as:

\begin{equation}
\label{sec:open_cross_entropy}
    \calL^{\text{open}}_{i} = - \frac{1}{K+1}\displaystyle \sum^{K+1}_{j=1} y_{i,j} \log \psi( C'(\kappa(\Phi(\bfX_{i}))))
\end{equation}
where $\psi(\cdot)$ is the softmax function for normalizing the network logits into a probability distribution. Note that in Eq.~(\ref{sec:open_cross_entropy}), for a target-private instance $\bfX^\ttarget_{i, s=1}$, we backpropagate only for the $(K+1)$\ths class.

\paragraph{\bf Overall objective} We train our model with a mini-batch using the following final objective:

\begin{equation}
    \calL = \mathcal{L}^{\text{open}} + \mathcal{L}^{\text{sup}} + \mathcal{L}^{\text{aug}} + \mathcal{L}^{\text{cross}} + \lambda \mathcal{L}^{\text{temp}}
    \label{eq:objective}
\end{equation}
where $\lambda$ is a weight for the temporal contrastive loss.

\noindent\textbf{Inference.} During inference the classifier $C'(\cdot)$ is used for classifying the target samples into one of the known or the unknown $(K+1)$\ths category.

\section{Experiments}
\label{sec:experiments}

\subsection{Experimental Set-up}



\paragraph{\bf Benchmarks} We evaluate our proposed framework under the \task task on four benchmarks that are derived from  UCF101~\citep{ucf}, HMDB~\citep{hmdb}, Olympic Sports~\citep{niebles2010modeling} and Epic Kitchens~\citep{epic_kitchens} action recognition datasets. The four benchmarks are described below.

\begin{table*}[!t]
    \centering
    \small
    \def\arraystretch{.7}
        \begin{tabular}{l|c|ccc|c|ccc|c}
            \toprule
            \multirow{2}{*}{Method} & \multirow{2}{*}{Backbone} & \multicolumn{4}{c|}{\textit{HMDB$\rightarrow$UCF}} & \multicolumn{4}{c}{\textit{UCF$\rightarrow$HMDB}} \\
            & & \textbf{ALL} & $\text{\bf OS}^{*}$ & \textbf{UNK} & \textbf{HOS} & \textbf{ALL} & $\text{\bf OS}^{*}$ & \textbf{UNK} & \textbf{HOS} \\
            \midrule
            DANN~\citep{dann} + OSVM~\citep{osvm} & \multirow{8}{*}{ResNet101} & 64.6 & 62.9 & 74.7 & 68.3 & 66.1 & 48.3 & 83.9 & 61.3 \\ 
            JAN~\citep{jan} + OSVM~\citep{osvm} & & 61.5 & 62.9 & 73.8 & 67.9 & 61.1 & 47.8 & 74.4 & 58.2 \\
            AdaBN~\citep{adabn} + OSVM~\citep{osvm} & & 62.9 & 58.8 & 73.3 & 65.3 & 62.9 & 58.8 & 73.3 & 65.3 \\
            MCD~\citep{mcd} + OSVM~\citep{osvm} & & 66.7 & 63.5 & 73.8 & 68.3 & 66.7 & 57.8 & 75.6 & 65.5 \\
            TA$^2$N~\citep{ta3n} + OSVM~\citep{osvm} & & 63.4 & 61.3 & 79.0 & 69.1 & 65.3 & 56.1 & 74.4 & 64.0 \\
            TA$^3$N~\citep{ta3n} + OSVM~\citep{osvm} & & 60.6 & 58.4 & 82.5 & 68.4 & 62.2 & 53.3 & 71.7 & 61.2 \\
            OSBP~\citep{Saito_2018_ECCV} + AvgPool & & 64.8 & 55.3 & 85.7 & 67.2 & 67.2 & 50.8 & 84.5 & 63.5 \\
            CEVT~\citep{cevt} & & 70.6 & 66.8 & 84.3 & 74.5 & 75.3 & 56.1 & 94.5 & 70.4 \\
            \midrule
            OVANet~\citep{ovanet} & \multirow{4}{*}{I3D} & \textbf{80.8} & \textbf{90.4} & 76.0 & 82.6 & 70.6 & 67.0 & 71.4 & 69.6 \\ 
            CEVT~\citep{cevt} & & 67.6 & 59.7 & \textbf{93.0} & 72.7 & 80.1 & 60.3 & \underline{96.1} & 74.1 \\
            \baseline (ours) & & \underline{79.3} & \underline{87.5} & 80.8 & \underline{84.0} & \underline{82.2} & \textbf{80.6} & 84.4 & \underline{82.5} \\
            \methodname (ours) & & 78.3 & 81.1 & \underline{88.7} & \textbf{84.7} & \textbf{89.1} & \underline{76.7} & \textbf{98.9} & \textbf{86.4} \\
            \bottomrule
        \end{tabular}
        \caption{Comparison with the state-of-the-art performance on the  \textit{HMDB$\leftrightarrow$UCF} benchmark. Best and second best numbers are highlighted in bold and underlines, respectively. The numbers of methods using ResNet101 as backbone are taken from \cite{cevt}. Overall, our proposed \methodname outperforms the state-of-the-art methods in terms of the important HOS metric}
    \label{tab:results_hu}
\end{table*}

\begin{table*}[!t]
    \centering
    \small
    \def\arraystretch{.6}
        \begin{tabular}{l|c|ccc|c|ccc|c}
            \toprule
            \multirow{2}{*}{Method} & \multirow{2}{*}{Backbone} & \multicolumn{4}{c|}{\textit{UCF$\rightarrow$Olympic}} & \multicolumn{4}{c}{\textit{Olympic$\rightarrow$UCF}} \\
            & & \textbf{ALL} & $\text{\bf OS}^{*}$ & \textbf{UNK} & \textbf{HOS} & \textbf{ALL} & $\text{\bf OS}^{*}$ & \textbf{UNK} & \textbf{HOS} \\
            \midrule
            DANN~\citep{dann} + OSVM~\citep{osvm} & \multirow{8}{*}{ResNet101} & 94.4 & 96.7 & 91.3 & 93.9 & 83.3 & 86.4 & 80.6 & 83.4 \\ 
            JAN~\citep{jan} + OSVM~\citep{osvm} & & 94.4 & {\bf 100.0} & 87.0 & 93.0 & 88.7 & 80.5 & \underline{95.5} & 87.3 \\
            AdaBN~\citep{adabn} + OSVM~\citep{osvm} & & 87.0 & 78.5 & {\bf 100.0} & 87.9 & 84.1 & 76.9 & 89.5 & 82.8 \\
            MCD~\citep{mcd} + OSVM~\citep{osvm} & & 87.0 & 86.7 & 87.0 & 86.8 & 83.7 & 85.5 & 82.1 & 83.8 \\
            TA$^2$N~\citep{ta3n} + OSVM~\citep{osvm} & & 96.3 & \textbf{100.0} & 91.3 & 95.4 & 87.9 & 78.5 & \underline{95.5} & 86.2 \\
            TA$^3$N~\citep{ta3n} + OSVM~\citep{osvm} & & 88.9 & 87.9 & 91.0 & 89.6 & 85.8 & 85.6 & 85.8 & 85.7 \\
            OSBP~\citep{Saito_2018_ECCV} + AvgPool & & 96.9 & 94.4 & \textbf{100.0} & 97.1 & 89.0 & 84.3 & 92.0 & 88.0 \\
            CEVT~\citep{cevt} & & \underline{98.1} & \underline{97.0} & \textbf{100.0} & \underline{98.5} & \underline{89.2} & 86.4 & 91.0 & 88.7 \\
            \midrule
            OVANet~\citep{ovanet} & \multirow{4}{*}{I3D} & 80.8 & 90.4 & 76.0 & 82.6 & 70.6 & 67.0 & 71.4 & 69.6 \\ 
            
            CEVT~\citep{cevt} & & 81.4 & 67.2 & \textbf{100.0} & 80.4 & 90.8 & 84.3 & \textbf{96.2} & \underline{89.9} \\
            
            \baseline (ours) &  & 92.8 & \textbf{100.0} & \textbf{100.0} & \textbf{100.0} & 84.1 & \textbf{95.8} & 76.4 & 85.0 \\
            \methodname (ours) & & \textbf{98.2} & \textbf{100.0} & \underline{95.0} & 97.4 & \textbf{90.8} & \underline{86.7} & 94.0 & \textbf{90.7} \\
            \bottomrule
        \end{tabular}
    \caption{Comparison to the state-of-the-art performance on the  \textit{UCF$\leftrightarrow$Olympic} benchmark. Overall, our proposed \methodname outperforms several state-of-the-art methods in terms of the HOS score. Best and second best numbers are highlighted in bold and underlines, respectively. The numbers of methods using ResNet101 as backbone are taken from \cite{cevt}. Note that this benchmark is smaller with only three shared and three unknown categories, and thus results have mostly saturated}
    \label{tab:results_uo}
\end{table*}

The \textit{HMDB$\leftrightarrow$UCF}~\citep{cevt} benchmark is constructed by collecting the 12 overlapping categories, viz., \texttt{Climb}, \texttt{Fencing}, \texttt{Golf}, \texttt{Kick Ball}, \texttt{Pull-up}, \texttt{Punch}, \texttt{Push-up}, \texttt{Ride Bike}, \texttt{Ride Horse}, \texttt{Shoot Ball}, \texttt{Shoot Bow} and \texttt{Walk}, present in the HMDB and UCF-101 datasets. Out of these 12 classes, the first six of them are selected as shared classes and the remaining six as target-private classes. Similarly, the \textit{UCF$\leftrightarrow$Olympic} benchmark contains six overlapping categories, viz., \texttt{Basketball}, \texttt{Clean and Jerk}, \texttt{Diving}, \texttt{Pole Vault}, \texttt{Tennis and Discus Throw}, from the UCF-101 and the Olympic Sports datasets, where the first half are shared classes and the rest are target-private ones. For both of these two benchmarks we follow the splits proposed in the CEVT~\citep{cevt}.

Given saturated performance on the above benchmarks, we also consider the benchmarks which is constructed from the egocentric action recognition dataset Epic Kitchens (EK). The original EK dataset is composed of first-person video sequences depicting egocentric actions (\eg ``\texttt{insert}'', ``\texttt{mix}'', \etc) performed in various kitchen environments, with a total of 32 environments. The \textit{closed-set} UVDA work MM-SADA~\cite{ek_multimodal} introduced a UVDA benchmark with three kitchens P01, P22 and P08 from the EK as the three domains D1, D2 and D3, respectively, with eight action classes (viz., \texttt{Put}, \texttt{Take}, \texttt{Open}, \texttt{Close}, \texttt{Wash}, \texttt{Cut}, \texttt{Mix}, \texttt{Pour}) overlapping among them. We adapt these three domains as a benchmark for the \task by additionally including all the instances from the non-overlapping classes of the target domain as target-private. 

\paragraph{\bf Evaluation metrics} To compare our proposed {\bf \methodname} to the baselines we adopt the metrics reported in the recent \task work \textbf{CEVT}~\citep{cevt}. Specifically, we report: the \textbf{ALL} (or the \textit{K}+1 way) accuracy that is the percentage of correctly predicted target samples over all the target samples; the $\text{\bf OS}^{*}$ is the averaged class accuracy over the \textit{known} classes only; the \textbf{UNK} recall metric which denotes the ratio of correctly predicted ``unknown'' target instances over the total number of ``unknown'' instances; and the \textbf{HOS} $=\frac{\text{OS}^* \times \text{UNK}}{\text{OS}^* + \text{UNK}}$ which is the harmonic mean between the known $\text{\bf OS}^{*}$ and unknown \textbf{UNK} accuracy. The \textbf{HOS} is the preferred metric in the open-set literature~\citep{hscore,ovanet} because the model can classify all the target instances to ``unknown'' class and attain 100\% \textbf{UNK} accuracy but 0\% $\text{\bf OS}^{*}$ accuracy. Thus, in order to have a high \textbf{HOS} score the model must do well in both the $\text{\bf OS}^{*}$ and the \textbf{UNK} metrics. Note that the \textbf{CEVT} also reports \textbf{OS}, the open-set average accuracy over the classes, but is similar to the \textbf{ALL} accuracy in nature. Being redundant, we do not report the \textbf{OS} accuracy.

\paragraph{\bf Implementation details} We use the I3D architecture as backbone, pretrained on Kinetics-400 as provided by \cite{kinetics_i3d}. We input \(c = 3\) clip-level features, each of dimension 1024 to the aggregation module $\kappa$, which outputs a 1024 dimensional feature $\bfh$. Both the projection head $g(\cdot)$ and $\kappa(\cdot)$ modules are implemented as a 2-layer MLP network. The classifiers \(C\) and \(C'\) are implemented as linear layers, with input 1024 (video-level feature) and output equal to \(K\) and \(K + 1\), respectively. In both the stages we trained our network with an SGD optimizer having momentum as 0.9, and learning rates 0.01 for the \textit{HMDB$\leftrightarrow$UCF} and \textit{UCF$\leftrightarrow$Olympic}, whereas 0.001 for the \textit{Epic-Kitchens}. We use batch size equal to 16 for all settings, and the parameter $\alpha$ was kept at the default value of 1. Our code is publicly available at \url{https://github.com/gzaraunitn/COLOSEO}.

\paragraph{\bf Baselines} We adopt the baselines from~\cite{cevt}, most of which are closed-set UDA methods adapted to the open-set scenario using {\bf OSVM}~\citep{osvm}. In details, we report {\bf DANN}~\citep{dann}, {\bf JAN}~\citep{jan}, {\bf AdaBN}~\citep{adabn}, {\bf MCD}~\citep{mcd}, {\bf OSBP}~\citep{Saito_2018_ECCV} and video-oriented {\bf TA$^2$N}, {\bf TA$^3$N}~\citep{ta3n} and {\bf CEVT}~\citep{cevt}. Moreover, we compare with an image-based OUDA method {\bf OVANet}~\citep{ovanet}, which we re-purposed to make it video-oriented. We additionally compare to the \textbf{CEVT}~\citep{cevt} that uses the I3D backbone~\citep{kinetics_i3d}, instead of the original Resnet-101, for a fair comparison. 

Finally, we also consider a variant of our proposed {\bf \methodname}, where the target-private rejection is carried out by thresholding the entropy computed by the closed-set classifier $C(\cdot)$, instead of using the source prototypes as described in Sec.~\ref{sec:method}. We call this variant as the {\bf \baseline}. Through our experiments we demonstrate that this entropy-based variant is often suboptimal with respect to our final method {\bf \methodname}.

\subsection{Comparison to the State of the Art}

\begin{table*}[t]
    \centering
    \small
    \def\arraystretch{0.9}
    \begin{tabular}{l|ccc|c|ccc|c|ccc|c}
        \toprule
        \multirow{2}{*}{\diagbox[width=4.9cm, height=.73cm]{Method $\downarrow$}{Setting $\rightarrow$}} & \multicolumn{4}{c|}{D2$\rightarrow$D1} & \multicolumn{4}{c|}{D3$\rightarrow$D1} & \multicolumn{4}{c}{D1$\rightarrow$D2} \\
        
        & \textbf{ALL} & $\text{\bf OS}^{*}$ & \textbf{UNK} & \textbf{HOS} & \textbf{ALL} & $\text{\bf OS}^{*}$ & \textbf{UNK} & \textbf{HOS} & \textbf{ALL} & $\text{\bf OS}^{*}$ & \textbf{UNK} & \textbf{HOS} \\
        \midrule
        CEVT~\citep{cevt} & 27.8 & 6.2 & \underline{65.7} & 11.3 & 32.3 & 4.2 & \textbf{93.6} & 8.0 & 18.9 & 5.4 & \underline{85.1} & 10.2 \\
        OVANet~\citep{ovanet} & 29.2 & 18.8 & 45.8 & 26.7 & \textbf{35.6} & \textbf{22.3} & 42.5 & \underline{29.3} & \underline{23.9} & 16.0 & 45.9 & 23.7 \\
        \baseline (ours) & \underline{36.3} & \underline{22.4} & \textbf{70.8} & \underline{34.0} & 32.4 & 11.7 & \underline{72.9} & 20.2 & 23.8 & \underline{17.0} & \textbf{88.2} & \underline{28.5} \\
        \methodname (ours) & \textbf{38.8} & \textbf{24.8} & 60.4 & \textbf{35.2} & \underline{33.1} & \underline{21.8} & 60.4 & \textbf{32.0} & \textbf{28.3} & \textbf{25.4} & 64.7 & \textbf{36.5} \\
        
        \midrule
        \multirow{2}{*}{\diagbox[width=4.9cm, height=.73cm]{Method $\downarrow$}{Setting $\rightarrow$}} & \multicolumn{4}{c|}{D3$\rightarrow$D2} & \multicolumn{4}{c|}{D1$\rightarrow$D3} & \multicolumn{4}{c}{D2$\rightarrow$D3} \\
        
        & \textbf{ALL} & $\text{\bf OS}^{*}$ & \textbf{UNK} & \textbf{HOS} & \textbf{ALL} & $\text{\bf OS}^{*}$ & \textbf{UNK} & \textbf{HOS} & \textbf{ALL} & $\text{\bf OS}^{*}$ & \textbf{UNK} & \textbf{HOS} \\
        \midrule
        CEVT~\citep{cevt} & 21.9 & 10.4 & \textbf{96.3} & 18.8 & 21.3 & 4.2 & \textbf{70.7} & 7.9 & 25.3 & 6.6 & \textbf{83.9} & 12.2 \\
        OVANet~\citep{ovanet} & \underline{30.8} & \underline{31.4} & 31.8 & 31.6 & \underline{28.5} & \underline{15.9} & 35.9 & \underline{22.0} & \textbf{34.0} & 16.7 & 50.0 & \underline{25.0} \\
        \baseline (ours) & 28.8 & 25.6 & \underline{70.5} & \underline{37.6} & 25.9 & 10.9 & \underline{70.3} & 18.9 & \underline{30.8} & \textbf{22.6} & \underline{56.2} & \textbf{32.2} \\
        \methodname (ours) & \textbf{33.3} & \textbf{34.3} & 47.0 & \textbf{39.7} & \textbf{34.4} & \textbf{24.9} & 39.0 & \textbf{30.4} & 32.1 & \underline{17.6} & 40.2 & 24.5 \\
        \bottomrule
    \end{tabular}
    \caption{Comparison to the state-of-the-art performance on the \textit{Epic-Kitchens} benchmark (with domains as D1, D2 and D3). All the methods use I3D as backbone. Best and second best numbers are highlighted in bold and underlines, respectively. Overall, our proposed \baseline and \methodname outperform the existing state-of-the-art methods in terms of the HOS score.}
    \label{tab:results_ek}
\end{table*}

We report in the Tab. \ref{tab:results_hu}, \ref{tab:results_uo} and \ref{tab:results_ek} the results obtained on the \textit{HMDB$\leftrightarrow$UCF}, \textit{UCF$\leftrightarrow$Olympic} and \textit{Epic-Kitchens} benchmarks, respectively. Overall, we observe from these tables that our proposed {\bf \methodname} outperforms the competitors in 8/10 adaptation settings in terms of the all important \textbf{HOS} metric.

In details, from the Tab. \ref{tab:results_hu} we can see that for the \textit{UCF$\rightarrow$HMDB} adaptation setting our proposed {\bf \methodname} yields the best \textbf{UNK} and \textbf{HOS} metrics, and the second best closed-set accuracy denoted by $\text{\bf OS}^{*}$. In addition, {\bf \methodname} achieves the best score for the \textbf{ALL} metric as well. On the other hand, for the reverse adaptation setting \textit{HMDB$\rightarrow$UCF}, although our {\bf \methodname} does not outperform the competitors in individual metrics, it surpasses them in terms of the \textbf{HOS}, by obtaining a better balance between $\text{\bf OS}^{*}$ and \textbf{UNK}. On the contrary, we can observe that several competitors fail to maintain a good trade-off between the $\text{\bf OS}^{*}$ and \textbf{UNK}, achieving a higher value in one or the other metric, but at the cost of a severe drop in the other one, resulting in a lower overall \textbf{HOS}.

As for \textit{UCF$\leftrightarrow$Olympic} benchmark, it is worth noting that the benchmark is very small (composed of only three known and three unknown classes) and has saturated performance. Nonetheless, we can observe in the Tab.~\ref{tab:results_uo} that {\bf \methodname} has competitive or better \textbf{HOS} scores when compared to its competitors. In particular, in the \textit{Olympic$\rightarrow$UCF} setting {\bf \methodname} achieves the best HOS score, whereas in the reverse direction our other variant {\bf \baseline} outperforms every method.

In the Tab.~\ref{tab:results_ek} we report the results for the challenging \textit{Epic-Kitchens} (EK) benchmark. For the EK benchmark we compare our proposed method with two existing state-of-the-art image-based \textbf{OVANet} and video-based \textbf{CEVT} methods. From a first glance we notice that the overall scores in the Tab.~\ref{tab:results_ek} is lower as compared to the \textit{HMDB$\leftrightarrow$UCF} and \textit{UCF$\leftrightarrow$Olympic} benchmarks. This can be attributed to: the challenging nature of the \textit{first-person} (egocentric) actions performed in cluttered indoor kitchen environments; and the mismatch of the I3D pre-training Kinetics dataset containing actions from \textit{third-person} views performed outdoor. Nevertheless, our two variants {\bf \baseline} and {\bf \methodname} outperforms the existing state-of-the-art in several adaptation settings of the EK. 

In particular, the results in the Tab.~\ref{tab:results_ek} show that {\bf \methodname}'s \textbf{HOS} score generally benefits from a higher closed-set accuracy $\text{\bf OS}^{*}$ when compared to other methods, although it doesn't achieve the best score with respect to the open-set accuracy \textbf{UNK}. Contrary to our proposed method, the competitor method \textbf{CEVT} is prone to classifying most of the target instances as ``unknown'', as evident from high \textbf{UNK} accuracy but very low closed-set accuracy $\text{\bf OS}^{*}$. This leads to the \textbf{CEVT} exhibiting significantly lower \textbf{HOS} scores when compared to our {\bf \baseline} and {\bf \methodname}.

\begin{table}[]
    \centering
    \def\arraystretch{.6}
    \resizebox{\columnwidth}{!}{
        \begin{tabular}{l c c c|c}
            \toprule
            Method & \textbf{ALL} & $\text{\bf OS}^{*}$ & \textbf{UNK} & \textbf{HOS} \\
            \midrule
            Ours w/o $\mathcal{L}^{\text{sup}}$ & 74.1 & 71.7 & 89.4 & 79.6 \\
            Ours w/o $\mathcal{L}^{\text{aug}}$ & 78.1 & \textbf{85.0} & 80.7 & 82.8 \\
            Ours w/o $\calL^{\text{cross}}$ & 74.8 & 73.5 & \textbf{90.0} & 80.9 \\
            Ours w/o $\mathcal{L}^{\text{temp}}$ & 74.1 & 79.6 & 77.1 & 78.4 \\
            Ours w/o $\mathcal{L}^{\text{sup}}$, $\mathcal{L}^{\text{aug}}$, $\calL^{\text{cross}}$ & 71.6 & 81.1 & 73.7 & 77.2 \\
            Ours (full) & \textbf{78.3} & 81.1 & 88.7 & \textbf{84.7} \\
            \bottomrule
        \end{tabular}
        }
    \caption{Ablation study of the contrastive losses in our \methodname in the \textit{HMDB$\rightarrow$UCF} adaptation setting}
    \label{tab:loss_ablation}
\end{table}

\subsection{Ablation Analysis}


We report the results of a thorough ablation analysis of the {\bf \methodname} framework conducted to better understand the impact of the proposed components and the sensitivity of {\bf \methodname} to the hyperparameters.

\paragraph{\bf Impact of the contrastive losses} First, we analyze the impact of the contrastive losses on the performance obtained by the proposed {\bf \methodname}. In  the Tab.~\ref{tab:loss_ablation} we report the performance of {\bf \methodname} for the adaptation setting \textit{HMDB$\rightarrow$UCF} when removing the individual contrastive losses described in Sec.~\ref{sec:method}. In details, we ablate by removing $\mathcal{L}^{\text{sup}}$, $\mathcal{L}^{\text{temp}}$, $\mathcal{L}^{\text{cross}}$ and $\mathcal{L}^{\text{aug}}$ one at a time from the final objective in Eq.~\ref{eq:objective}. We can observe that all the contrastive losses positively contribute to the final score and removing any of them negatively impacts the metrics. Notably, the absence of the newly proposed temporal contrastive loss $\calL^{\text{temp}}$ leads to a significant drop in performance (by -6.3\% points). This highlights the importance of leveraging the temporal information in videos for a well structured representation space, which can be realized for free without needing any labels.

Furthermore, note that even without the cross-domain alignment loss $\calL^{\text{cross}}$ the performance is relatively high, with a drop of only 3.8\% points, which is lower than without the $\calL^{\text{temp}}$ loss (6.3\% points drop). It hints at the fact that the $\calL^{\text{temp}}$ and $\calL^{\text{aug}}$ can effectively and implicitly align the two domains without needing any alignment technique.


\begin{table}[!h]
    \centering
    \small
    \resizebox{\columnwidth}{!}{
        \begin{tabular}{lccc|c|ccc|c}
            \toprule
            \multirow{2}{*}{Method} & \multicolumn{4}{c|}{\# clips = 3} & \multicolumn{4}{c}{\# clips = 4} \\
             & \textbf{ALL} & $\text{\bf OS}^{*}$ & \textbf{UNK} & \textbf{HOS} & \textbf{ALL} & $\text{\bf OS}^{*}$ & \textbf{UNK} & \textbf{HOS} \\
             \midrule
             COP & 72.5 & 73.4 & \textbf{90.0} & 80.8 & 53.8 & \textbf{73.6} & 42.1 & 53.5 \\ 
             \methodname & \textbf{78.3} & \textbf{81.1} & 88.7 & \textbf{84.7} & \textbf{67.9} & 63.1 & \textbf{84.7} & \textbf{72.3} \\ 
             \midrule
        \end{tabular}
    }
    \caption{Comparison of the \methodname to the COP baseline and the impact of the number of clips $c$ in the \textit{HMDB$\rightarrow$UCF} adaptation setting}
    \label{tab:cop_ablation}
\end{table}

\begin{table*}[h!]
    \centering
    \small
    \def\arraystretch{.8}
    \begin{tabular}{l|ccc|c|ccc|c}
        \toprule
        \multirow{2}{*}{Method} & \multicolumn{4}{c|}{\textit{HMDB}$\rightarrow$\textit{UCF}} & \multicolumn{4}{c}{\textit{UCF}$\rightarrow$\textit{HMDB}} \\
            & \textbf{ALL} & $\text{\bf OS}^{*}$ & \textbf{UNK} & \textbf{HOS} & \textbf{ALL} & $\text{\bf OS}^{*}$ & \textbf{UNK} & \textbf{HOS} \\
            \midrule
            CEVT~\citep{cevt}
            & 66.7$\pm$0.9 & 49.9$\pm$9.2 & 87.2$\pm$6.4 & 62.8$\pm$8.0 & 72.6$\pm$5.7 & 57.7$\pm$11.6 & 89.2$\pm$5.9 & 64.1$\pm$7.9 \\
        
        \baseline (ours)
            & 83.1$\pm$3.8 & 79.3$\pm$7.4 & 87.8$\pm$2.5 & 83.0$\pm$2.8 & 81.9$\pm$0.4 & 74.4$\pm$6.3 & \textbf{94.8}$\pm$7.3 & \textbf{82.9}$\pm$3.1 \\
        \methodname (ours)
            & \textbf{83.7}$\pm$3.8 & \textbf{80.1}$\pm$0.8 & \textbf{90.9}$\pm$2.5 & \textbf{85.4}$\pm$0.5 & \textbf{83.5}$\pm$3.9 & \textbf{74.9}$\pm$4.2 & 93.0$\pm$4.6 & \textbf{82.9}$\pm$2.9 \\
        \bottomrule
    \end{tabular}
    \caption{Comparison of the averaged performance under three different splits of the shared and unknown classes on the \textit{HMDB$\leftrightarrow$UCF} benchmark. Each entry reports the average and the standard deviation of a metric across three splits. All the methods use I3D as backbone. Best average performance is highlighted in bold. Overall, the \methodname consistently and substantially outperforms the CEVT, while having lower variance than CEVT in both the adaptation settings.
    }
    \label{tab:hu_splits}
\end{table*}

\paragraph{\bf Comparison to video-based pretext task} 
Temporal information offered by video data has previously been exploited in a UVDA method \textbf{SAVA}~\citep{sava} that in particular uses the clip ordering prediction (COP) as a pretext task. In details, the COP consist in training the model to predict the correct order in which the input clips have been permuted. The key idea is that while trying to distinguish between the input clips that are temporally shuffled from the non-shuffled (or original) ones, the network will learn salient and representative features from the actions. Given that our proposed temporal loss $\mathcal{L}^{\text{temp}}$ shares similar spirit with the pretext task of COP, we also compare and report in Tab.~\ref{tab:cop_ablation} the results obtained with {\bf \methodname}, except by replacing $\mathcal{L}^{\text{temp}}$ with the COP loss, for $c=3$ and $c=4$ clips. It is evident from the Tab.~\ref{tab:cop_ablation} that our $\mathcal{L}^\text{temp}$ significantly outperforms the \textbf{COP} baseline, which produces much more unstable results, with a tendency to either over-accept or over-reject samples. Furthermore, when using 4 clips, the COP loss produces a drastic drop with respect to the final score, as the ordering problem becomes more challenging (being a $c!$-way classification problem), which aligns with the findings in \cite{sava}. In contrast to the COP loss, our temporal loss is significantly more robust to the increased number of clips, which is evident by an increased gap in the \textbf{HOS} scores between the two methods.

\begin{figure}
    \centering
    \includegraphics[width=0.8\columnwidth]{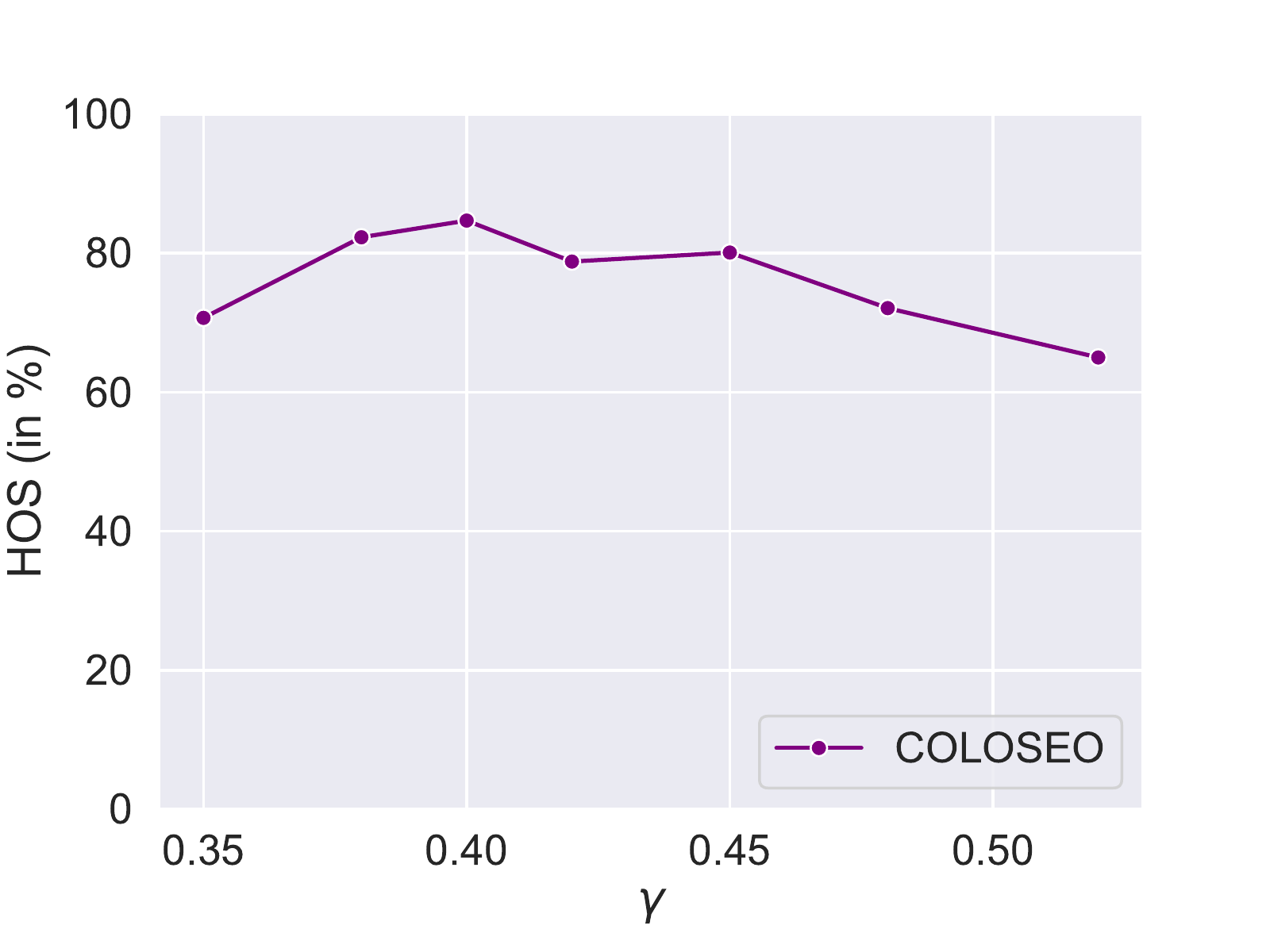}
    \caption{Impact on the \textbf{HOS} score while varying the threshold value $\gamma$, which is used for the known/unknown target separation in the \methodname, in the \textit{HMDB$\rightarrow$UCF adaptation setting}
    }
    \label{fig:sensitivity_study}
\end{figure}

\paragraph{\bf Sensitivity to hyperparameters} In the Fig.~\ref{fig:sensitivity_study} we demonstrate the sensitivity analysis on the hyperparameter $\gamma$ described in the Eq.~(\ref{eqn:open-set-separation}). The $\gamma$ governs whether to assign a target instance to a shared class or reject it as target-private class. The Fig.~\ref{fig:sensitivity_study} shows that \textbf{HOS} score of our {\bf \methodname} remains fairly stable over a wide range of $\gamma$ values. From our experiments we observe that extreme low or high values of $\gamma$ results in the protocol over-accepting and over-rejecting target instances, respectively, thus favoring one of the metrics between $\text{\bf OS}^{*}$ and \textbf{UNK} but impacting negatively the final \textbf{HOS} score. Thus, we limit the values of $\gamma$ within reasonable limits despite the theoretical bounds of $\gamma$ lies in [0, 1]. 

\begin{figure}
    \centering
    \includegraphics[width=0.8\columnwidth]{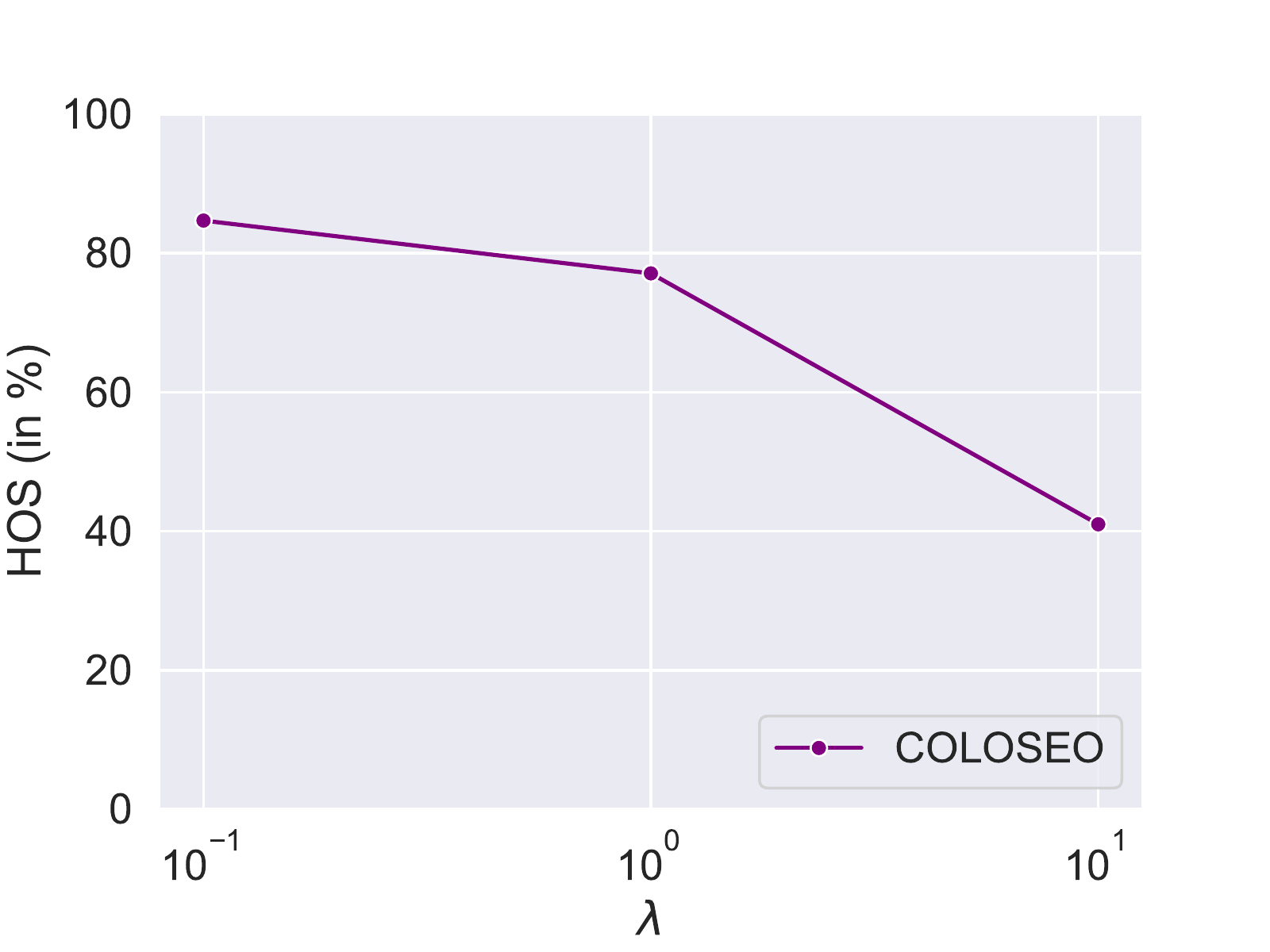}
    \caption{Impact on the \textbf{HOS} score while varying $\lambda$, which weights the temporal contrastive loss, in the \textit{HMDB$\rightarrow$UCF} adaptation setting. Note that the x-axis is in logarithmic scale}
    \label{fig:sensitivity_study2}
\end{figure}

We also ablate on the second hyperparameter $\lambda$ that controls the contribution of the temporal contrastive loss $\mathcal{L}^\text{temp}$ to the final objective described in Eq.~\ref{eq:objective}. Given our implementation of the $\mathcal{L}^\text{temp}$, described in Eq.~\ref{eq:triplet_loss}, uses a triplet loss instead of the traditional InfoNCE contrastive objective~\citep{gutmann2010noise}, we had to balance the maginitude of the $\mathcal{L}^\text{temp}$. In the Fig.~\ref{fig:sensitivity_study2} we plot the \textbf{HOS} score by varying the $\lambda$ in \textit{logarithmic} scale. We can observe that the chosen value in the final version of our framework (i.e. $\lambda=$ 0.1) leads to the best \textbf{HOS} score. Furthermore, it emerges that the \textbf{HOS} score remains quite stable until $\lambda=1$, and only degrades for very large values ($\lambda=10$), which is a reasonable behaviour.



\paragraph{\bf Impact of shared/unknown classes splits} As shown in the open-set recognition literature~\citep{vaze2021open}, the \textbf{UNK} performance relies heavily on the split between the shared and OOD classes, it also becomes imperative to test the \task methods under different shared/unknown classes splits. To this end, we report in Tab.~\ref{tab:hu_splits} the averaged results obtained by our proposed method and compare them to the best competitor \textbf{CEVT} under three random splits of 6 shared and 6 unknown classes of the \textit{HMDB$\leftrightarrow$UCF} benchmark. We can see that our {\bf \methodname} and {\bf \baseline} outperforms the \textbf{CEVT} by significant margins even under different different splits of the shared and unknown classes. More interestingly, the standard deviation in the \textbf{HOS} score attained by our {\bf \methodname} is also substantially lower than that of \textbf{CEVT} (0.5 versus 8.0 and 2.9 versus 7.9 in the \textit{HMDB$\rightarrow$UCF} and \textit{UCF$\rightarrow$HMDB} adaptation settings, respectively). This shows that irrespective of the shared action categories present in the source, the representation space learned by {\bf \methodname} is more compact and thereby rejecting unknown classes becomes easier than the weighted adversarial strategy used in the \textbf{CEVT}.


\begin{figure}[h!]
    \centering
    \includegraphics[width=0.8\columnwidth]{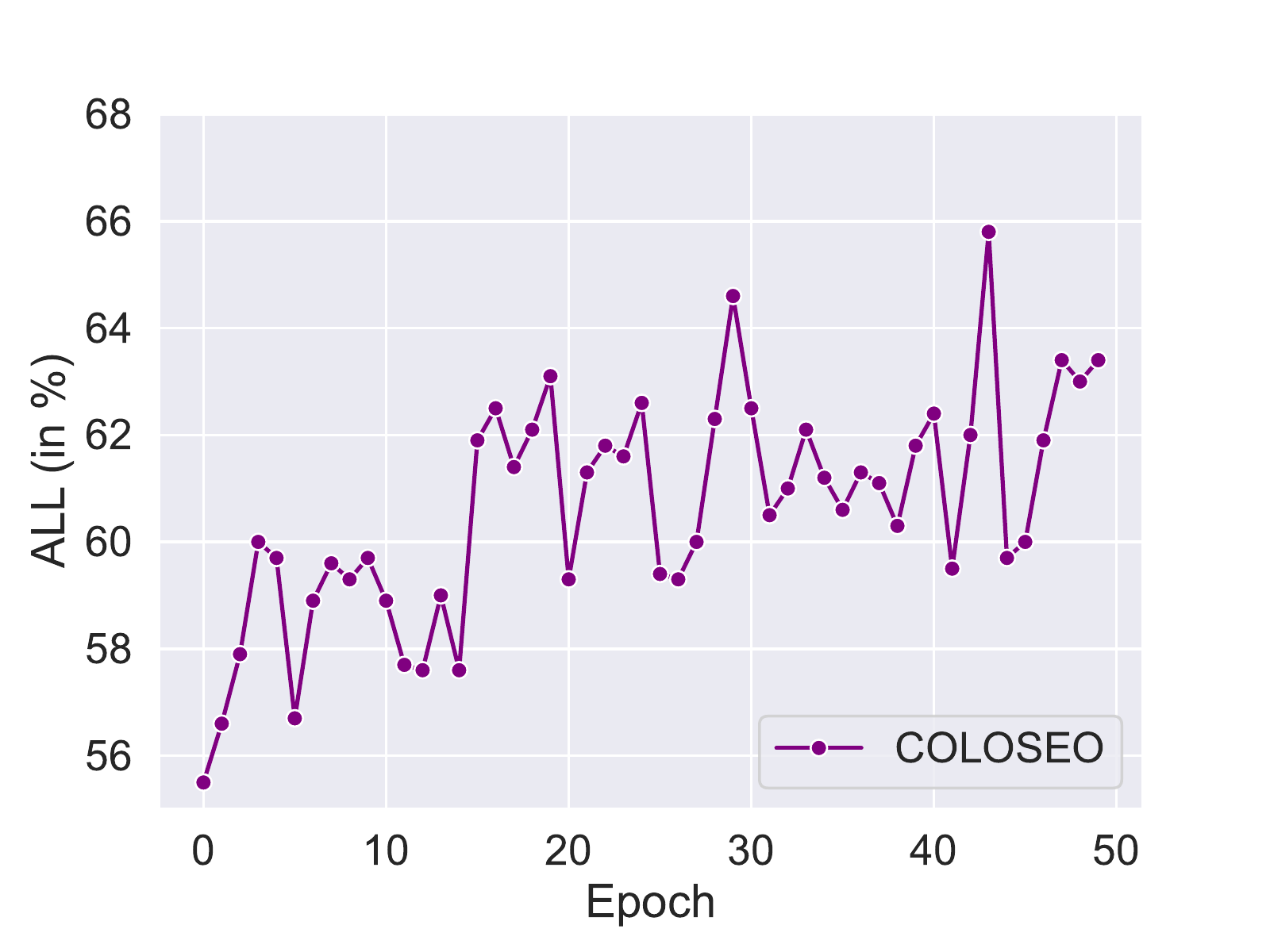}
    \caption{The evolution of the pseudo-label \textbf{ALL} accuracy of \methodname on the target domain with the progression of training in the \textit{HMDB$\rightarrow$UCF} adaptation setting}
    \label{fig:pseudo}
\end{figure}

\paragraph{\bf Quality of pseudo-labels} Given our cross-domain contrastive loss $\calL^{\text{cross}}$ in the Eq.~\ref{eq:loss_cross_domain} relies on the pseudo-labels computed on the target, in Fig.~\ref{fig:pseudo} we visualize the \textbf{ALL} accuracy metric of the pseudo-labels for the \textit{HMDB$\rightarrow$UCF} adaptation setting, as the training progresses. As expected, we observe that \textbf{ALL} pseudo-label accuracy on the target training data is increasing as the training progresses. This in turn positively impacts the $\calL^{\text{cross}}$ to better align the source and the shared target classes. Moreover, as the \textbf{ALL} also includes the accuracy of the unknown class prediction, it also indirectly reflects the usefulness of the unknown target rejection. 



\section{Conclusions}
\label{sec:conclusion}

In this work we presented the {\bf \methodname} framework for addressing open-set unsupervised video domain adaptation. Our proposed {\bf \methodname} leverages contrastive learning under complementary yet unified formulations to: (i) learn discriminative and compact feature representations in both the source and target domains; and (ii) align source and target distributions with respect to the shared classes. In particular, the {\bf \methodname} includes a video oriented temporal contrastive loss that clusters different actions by exploiting the temporal information available freely in video data. We showed that learning compact representations can simplify the separation between the shared and unknown classes. We carried out an extensive experimental evaluation on three \task benchmarks and demonstrated that our {\bf \methodname} significantly outperforms the existing state-of-the-art methods.

\bibliographystyle{model2-names}
\bibliography{main}

\end{document}